\documentclass[conference]{IEEEtran}
\usepackage{amsmath,graphicx}
\usepackage{mathtools}
\usepackage{xcolor}
\usepackage{amsfonts}
\usepackage{xfrac}
\usepackage{amssymb}
\usepackage{multirow}
\usepackage{varwidth}
\usepackage{algorithm,algorithmicx}
\usepackage{algpseudocode}
\usepackage{booktabs}   
\usepackage{cite}
\usepackage{amsmath,amssymb,amsfonts}
\usepackage{graphicx}
\usepackage{textcomp}
\usepackage{xcolor}
\usepackage{fancyhdr}

\fancypagestyle{alim}{\fancyhf{}\fancyfoot[L]{\footnotesize \textit{The paper has been accepted at the 37th IEEE International Conference on Tools with Artificial Intelligence (ICTAI) 2025}}}

\begin{document}

\title{
SCALAR: Self-Calibrating Adaptive Latent Attention Representation Learning}


\author{%
    \begin{minipage}{1.0\textwidth}
        \centering
        Farwa Abbas$^{\ast}$,
        Hussain Ahmad$^{\dagger}$,
        Claudia Szabo$^{\dagger}$ \\
        \smallskip
        $^{\ast}$Imperial College London, UK;
        $^{\dagger}$The University of Adelaide, Australia \\
        f.abbas20@imperial.ac.uk;        hussain.ahmad@adelaide.edu.au; claudia.szabo@adelaide.edu.au
\vspace{-1.0\baselineskip}
    \end{minipage}
}


\maketitle

\begin{abstract}
High-dimensional, heterogeneous data with complex feature interactions pose significant challenges for traditional predictive modeling approaches. While Projection to Latent Structures (PLS) remains a popular technique, it struggles to model complex non-linear relationships, especially in multivariate systems with high-dimensional correlation structures. This challenge is further compounded by simultaneous interactions across multiple scales, where local processing fails to capture cross-group dependencies. Additionally, static feature weighting limits adaptability to contextual variations, as it ignores sample-specific relevance. To address these limitations, we propose a novel method that enhances predictive performance through novel architectural innovations. Our architecture introduces an adaptive kernel-based attention mechanism that processes distinct feature groups separately before integration, enabling capture of local patterns while preserving global relationships. 
Experimental results show substantial improvements in performance metrics, compared to the state-of-the-art methods across diverse datasets.
\end{abstract}

\begin{IEEEkeywords}
Latent representation learning, Attention network, Kernel, Variational encoding, Drug-target interactions
\end{IEEEkeywords}

\section{Introduction}

Effectively modeling high-dimensional data with complex, heterogeneous features remains a persistent challenge in scientific and industrial domains \cite{jia2022feature}. Datasets in applications such as genomics, finance, cyber-security and pharmacology often exhibit non-linear dependencies, sparse and imbalanced class distributions, and entangled feature relationships \cite{montesinos2022overfitting, vogelstein2021supervised, goel2024machine, chopra2024chatnvd, ahmad2025survey, ullah2025skills,ahmad2025future}. These characteristics amplify issues like overfitting, reduced generalization, and high computational overhead, particularly in scenarios where feature dimensionality exceeds the number of samples.

\thispagestyle{alim}
To address these limitations, dimensionality reduction techniques have been widely explored. Linear methods such as Principal Component Analysis (PCA) and Partial Least Squares (PLS) provide interpretable transformations but often fail to capture non-linear feature dependencies. Kernel-based extensions like Kernel PLS (KPLS) \cite{rosipal2001kernel} attempt to overcome this by leveraging non-linear embeddings at the cost of scalability. Manifold learning methods such as t-SNE and UMAP \cite{mcinnes2020umapuniformmanifoldapproximation} offer superior preservation of local structures but provide embeddings that are difficult to interpret in terms of global structure or distances \cite{kobak2021initialization}, limiting their utility for predictive tasks. Meanwhile, autoencoders provide flexible non-linear mappings but often degenerate to identity functions without appropriate regularization \cite{bengio2013representation}.

More recently, attention-based architectures have gained prominence for their ability to dynamically weight informative features, providing an adaptable alternative to fixed transformations. These methods have shown promise in structured data domains but remain challenged by training instability, high computational cost, and poor handling of heterogeneous feature groups \cite{xu2019understanding, abbas2024robust, arik2021tabnet, jain2019attention}. These limitations are particularly pronounced in biomedical applications, where data is often high-dimensional, noisy, and derived from diverse modalities. In this context, drug-target interaction (DTI) prediction serves as a representative problem that exemplifies the complexities of learning from high-dimensional, complex biological data. DTI aims to identify potential binding relationships between chemical compounds and target proteins. Accurate DTI modeling accelerates the identification of bioactive compounds, reducing the cost and time of experimental screening. However, DTI datasets are inherently multimodal, comprising heterogeneous features derived from diverse sources such as molecular fingerprints, graph embeddings, and protein sequences. These diverse representations pose significant challenges for feature integration, generalization under limited data, and interpretability.

To address these challenges, we propose a Self-Calibrating Adaptive Latent Attention Representation (SCALAR) network, a novel architecture designed for learning from high-dimensional, multimodal data. It employs a hierarchical structure with adaptive attention mechanisms that selectively capture both intra- and inter-group feature dependencies, enabling more effective representation learning across heterogeneous inputs. Additionally, it integrates variational regularization to promote generalization and stability in data-scarce settings. Our key contributions are summarized as follows:

\begin{enumerate}
\item We implement a hierarchical architecture that processes distinct feature groups separately before global integration, specifically addressing the challenge of information loss during feature fusion in heterogeneous datasets. Experimental validation shows this approach maintains biological relevance in feature selection (identifying clinically relevant kinases like EGFR and RPS6KA5) while achieving 61\% error reduction compared to methods that process features uniformly.

\item We introduce an adaptive regularization mechanism that dynamically adjusts dropout rates and scaling factors based on input characteristics, directly addressing the overfitting challenge in high-dimensional, low-sample scenarios. 

\item  We incorporate KL divergence regularization within a variational framework to prevent representational collapse and improve model stability. This addresses the training instability issues commonly observed in attention-based methods for tabular data, as evidenced by consistent performance across diverse datasets without extensive hyperparameter tuning.
\end{enumerate}

Extensive experiments on benchmark DTI and NIR spectroscopy datasets show that SCALAR outperforms both traditional PLS methods and recent attention-based models in predictive performance, with SHAP analyses offering insight into feature contributions.

\section{Related Work}

Partial Least Squares (PLS) regression has been widely applied to high-dimensional datasets, particularly in the presence of multicollinearity. PLS is widely used in fields like chemometrics \cite{kumar2021partial}, metabolomics \cite{anwardeen2023statistical}, and neuroimaging \cite{mihalik2022canonical}, where it projects high-dimensional features into a latent space designed to maximize their correlation with output variables. However, its reliance on linear assumptions limits its applicability in complex, nonlinear settings.

Kernel PLS (KPLS) \cite{rosipal2001kernel} addresses this limitation through the use of kernel functions but incurs high computational costs due to its $O(n^2)$ memory complexity. More efficient variants, such as Improved Kernel PLS (IKPLS) \cite{engstrom2024ikpls}, mitigate these costs but face challenges in kernel selection especially when dealing with heterogeneous data \cite{briscik2024supervised}. Deep learning extensions to PLS \cite{kong2022deep} offer greater flexibility but often suffer from overfitting and require large datasets for stable training.

Unsupervised methods like autoencoders and manifold learning techniques (e.g., t-SNE and UMAP) have also been explored for representation learning. While these methods excel at preserving local structure \cite{kobak2021initialization}, their stochastic nature and lack of inverse mapping limit interpretability and integration into downstream predictive models. In contrast, attention mechanisms provide dynamic feature weighting and improved adaptability to complex dependencies \cite{xu2019understanding, jain2019attention}. Yet, they remain sensitive to training instability and often require extensive regularization, particularly in structured or tabular domains \cite{arik2021tabnet}.

These limitations are especially pronounced in domains like biomedical informatics, where data is high-dimensional, structured, and multimodal. One representative task is drug–target interaction (DTI) prediction, which seeks to identify binding affinities or interaction likelihoods between small molecules and protein targets, addressing key challenges in computational drug discovery. Early models such as DeepDTA \cite{ozturk2018deepdta} employed convolutional neural networks to encode drug and protein representations independently, followed by concatenation for prediction. While effective, these architectures assumed uniform feature importance and lacked explicit mechanisms for modeling cross-modal interactions.

Subsequent models, including DeepConv-DTI \cite{lee2019deepconv} and DLM-DTI \cite{lee2024dlm}, introduced architectural refinements but continued to fall short in capturing fine-grained compound–protein dependencies.
More recently, attention mechanisms have been explored to better model protein–ligand binding \cite{proteinLigandBinding2024, 10849366}, highlighting their potential to capture complex dependencies.

In parallel, Attention-PLS \cite{xiong2022soft} incorporated attention mechanisms into the PLS framework to improve interaction modeling. While promising, this approach was not specifically designed to handle heterogeneous feature groups, highlighting the need for more flexible methods that can effectively integrate diverse data modalities. Other adaptive architectures, such as Squeeze-and-Excitation networks \cite{hu2018squeeze} and dynamic convolutions \cite{chen2020dynamic}, have shown success in vision tasks and have been adapted to structured data in models like WTSN \cite{kadra2021well} and AutoInt \cite{song2019autoint}. However, many of these methods remain sensitive to data sparsity and require extensive regularization.

Hierarchical and probabilistic modeling techniques offer complementary benefits. Ladder Variational Autoencoders (LVAEs) \cite{sonderby2016ladder} introduce hierarchical latent spaces for greater expressiveness, while Bayesian PLS \cite{vidaurre2013bayesian} captures uncertainty in latent representations. Yet, integrating these approaches with modern attention-based architectures remains largely unexplored. To address these challenges, we propose SCALAR as a unified attention-based framework that integrates hierarchical feature processing, self-calibration, and variational learning. This approach enables more accurate representation of complex, multimodal data while preserving model stability and adaptability, making SCALAR an effective tool for regression tasks that involve complex feature interactions. In the subsequent section, we provide a detailed formulation of the problem.

\section{Problem Formulation}

\begin{figure*}[t]
    \centering
    \includegraphics[width=0.8\linewidth]{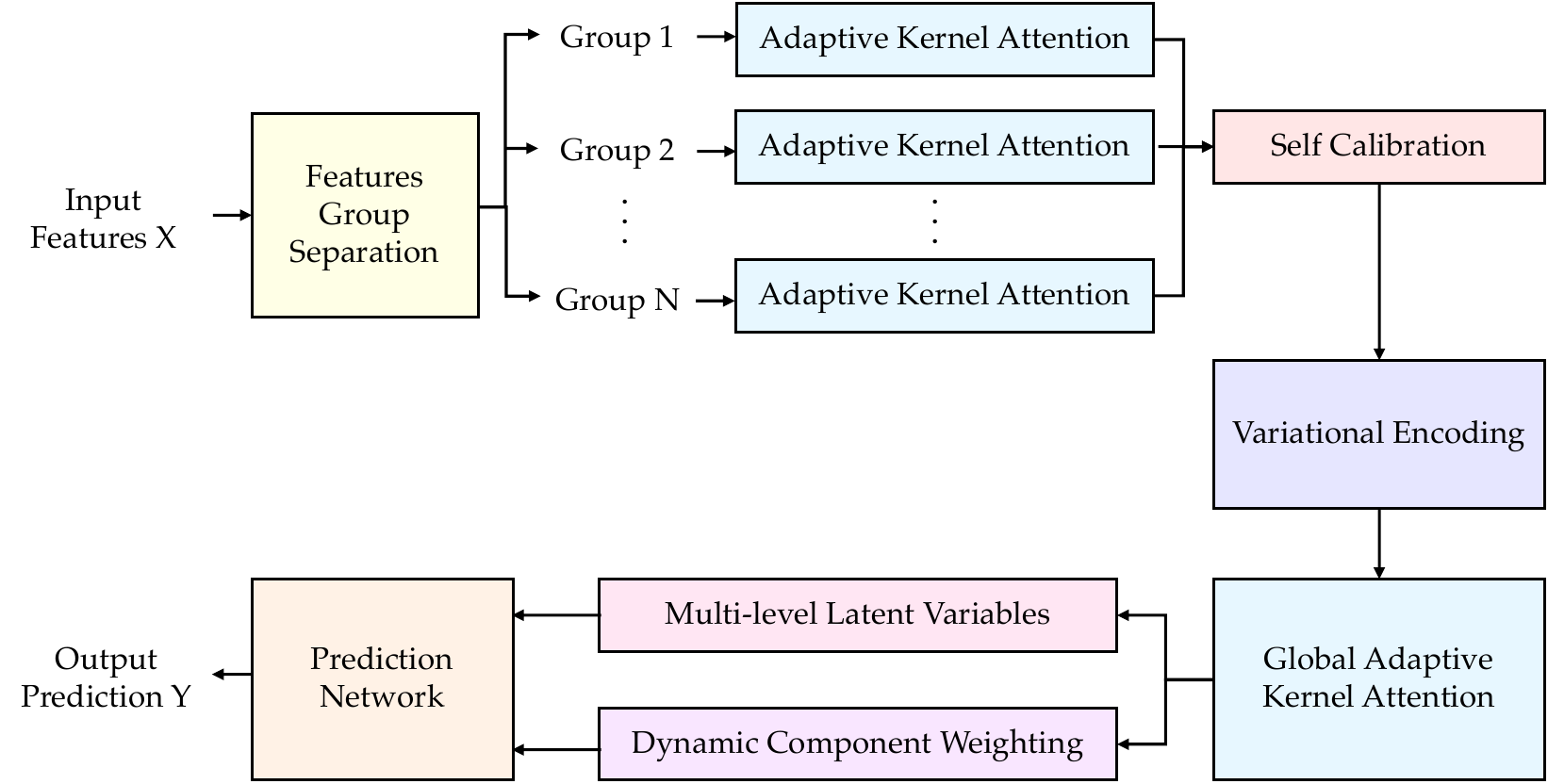}
    \caption{Block diagram of our proposed SCALAR model}
    \label{block_diag}
    \vspace{-1\baselineskip}
\end{figure*}

In this section, we present a mathematical formulation of the problem statement. Let $\mathbf{X} \in \mathbb{R}^{n \times p}$ be the input data matrix with $n$ samples and $p$ features, and $\mathbf{y} \in \mathbb{R}^{n \times 1}$ be the corresponding response variable. We assume that the features in $\mathbf{X}$ can be divided into $G$ groups, where each group $g$ contains features with indexes from $s_g$ to $e_g$. The overall process is illustrated in the block diagram shown in Figure \ref{block_diag}. Formally, we define the feature matrix for group $g$ as:
\begin{align}
   \mathbf{X}^{(g)} = \mathbf{X}_{:, s_g:e_g} \in \mathbb{R}^{n \times p_g}, 
\end{align}
where $p_g = e_g - s_g$ is the number of features in group $g$.
Our goal is to learn a model that effectively captures the complex, non-linear relationships between $\mathbf{X}$ and $\mathbf{y}$ while accommodating the heterogeneous nature of different feature groups. For each feature group $g$, we apply an adaptive kernel attention mechanism to capture important feature interactions within the group. Let $\mathbf{X}^{(g)}_i \in \mathbb{R}^{p_g}$ be the feature vector of the $i$-th sample in group $g$. 
Given that features often exhibit localized dependencies within specific groups, applying a global transformation may obscure meaningful interactions. 
Therefore, we define group-specific feature sets and construct data-dependent adaptive kernels followed by normalization to ensure numerical stability and prevent any particular feature dimension from dominating the representation. 
\begin{align}
   \mathbf{K}^{(g)}_i = \phi_K(\mathbf{X}^{(g)}_i) \in \mathbb{R}^{k \times p_g}, \hspace{0.5cm}     \hat{\mathbf{K}}^{(g)}_{i,j} = \frac{\mathbf{K}^{(g)}_{i,j}}{\|\mathbf{K}^{(g)}_{i,j}\|_2} \in \mathbb{R}^{p_g},
\end{align}
where $\phi_K$ is a non-linear function implemented as a neural network with parameters $\theta_K$, and $k$ is the number of kernels, and $\hat{\mathbf{K}}^{(g)}_{i,j} \in \mathbb{R}^{p_g}$ is the $j$-th normalized kernel for sample $i$.
The normalized kernels are used to transform the original features, selectively modulating feature dimensions based on their relevance as indicated by the learned kernels. We then compute a weighted sum of these kernel-transformed features:
\begin{align}\hat{\mathbf{w}}^{(g)}_i &= \text{softmax}(\phi_W(\mathbf{X}^{(g)}_i)), \\\mathbf{A}^{(g)}_i &= \sum_{j=1}^{k} \hat{\mathbf{w}}^{(g)}_{i,j} (\mathbf{X}^{(g)}_i \odot \hat{\mathbf{K}}^{(g)}_{i,j} ), \hspace{0.2cm}.\end{align}
The kernel weights $\hat{\mathbf{w}}^{(g)}$ are obtained using a softmax function, which ensures that the weights sum to 1 and provides a probabilistic interpretation of each kernel’s importance for every sample. $\phi_W$ is another non-linear function with parameters $\theta_W$. 
We apply a projection with a residual connection to preserve gradient flow and maintain access to the original feature information and after processing each feature group separately, we combine the attended features inside $\mathbf{Z}_i$ as:
\begin{align}\mathbf{Z}^{(g)}_i &= \phi_P(\mathbf{A}^{(g)}_i) + \mathbf{X}^{(g)}_i \in \mathbb{R}^{p_g}, \\\hspace{0.5cm} \mathbf{Z}_i &= [\mathbf{Z}^{(1)}_i, \mathbf{Z}^{(2)}_i, \ldots, \mathbf{Z}^{(G)}_i] \in \mathbb{R}^{p},\end{align}
where $\phi_P$ is a linear projection with parameters $\theta_P$. 
This hierarchical grouping and kernel adaptation methodology ensures that local feature interactions are preserved before global feature integration, ultimately leading to more expressive representations that capture both group-specific and cross-group dependencies.

\subsubsection{Self-Calibration and Variational Processing}

The self-calibration layer is designed to adaptively adjust feature representations based on the characteristics of each input sample. It learns a set of scaling factors $\gamma_i$ that modulate the importance of transformed features, allowing the model to emphasize more informative components while down-weighting less relevant ones. After transformation, these scaled features are combined with the original input through a residual connection. This not only preserves the original information but also stabilizes learning by allowing the model to fall back on the unaltered input when needed. This is particularly useful to enhance flexibility and robustness, in settings with heterogeneous or noisy data.
\begin{align}
\mathbf{S}_i = \mathbf{Z}_i + \gamma_i (\phi_T(\mathbf{Z}_i) \odot \mathbf{M}_i / (1 - \delta_i) ),\end{align}
where $\phi_T$ is a non-linear transformation with parameters $\theta_T$, and $\mathbf{M}_i \in \{0, 1\}^{p}$ is a random binary mask in which each element is 1 with probability $1 - \delta_i$. The dropout rate $\delta_i$ and scaling factor $\gamma_i$ are computed as $[\delta_i, \gamma_i] = \phi_C(\mathbf{Z}_i) \in \mathbb{R}^{2}$, where $\phi_C$ is another non-linear function with parameters $\theta_C$, $\delta_i \in [0, 0.4]$, and $\gamma_i \in [0.5, 1.0]$. The normalization factor $(1 - \delta_i)^{-1}$ ensures the expected value remains stable despite dropout, supporting consistent training dynamics.

To further enhance representational robustness, we integrate a principled variational framework that encodes transformed features into a probabilistic latent distribution. 
This intermediate representation is then projected onto distribution parameters characterizing a multivariate Gaussian latent space:
\begin{align*}\boldsymbol{\mu}_i = \phi_{\mu}(\phi_E(\mathbf{S}_i)) \in \mathbb{R}^{d}, \hspace{0.3cm}\log \boldsymbol{\sigma}_i = \phi_{\sigma}(\phi_E(\mathbf{S}_i)) \in \mathbb{R}^{d},\end{align*}
where  hidden layer $\phi_E(\mathbf{S}_i)\in \mathbb{R}^{p/2}$ and $\phi_E, \phi_{\mu}, \phi_{\sigma}
$ represent non-linear functions with parameters $\theta_E, \theta_{\mu}$, and $\theta_{\sigma}$ respectively, and $d$ denotes the dimensionality of the latent space.
To facilitate gradient-based optimization through the sampling process, we employ the reparameterization technique and the sampled latent variable is subsequently decoded back to the original feature space as:
\begin{align}
\mathbf{z}_i &= \boldsymbol{\mu}_i + \boldsymbol{\epsilon}_i \odot \exp(\log \boldsymbol{\sigma}_i / 2) \in \mathbb{R}^{d}, \hspace{0.5cm} \\\mathbf{V}_i &= \mathbf{S}_i + \phi_D(\mathbf{z}_i) \in \mathbb{R}^{p},
\label{eq10}
\end{align}
where $\boldsymbol{\epsilon}_i \sim \mathcal{N}(0, \mathbf{I})$ represents a standard normal noise vector, and $\phi_D$ represents a non-linear function with parameters $\theta_D$. This allows gradients to pass through the stochastic sampling step, enabling efficient and stable optimization. By combining self-calibrated features with the original input through a residual connection, the model retains useful raw information while reinforcing important learned patterns. To prevent overfitting and encourage well-structured latent representations, we apply variational regularization using the Kullback–Leibler (KL) divergence, which pulls the learned distribution closer to a standard normal prior:
\begin{align}
\mathcal{L}_{\text{KL}} = \frac{1}{n} \sum_{i=1}^{n} \frac{1}{2} \sum_{j=1}^{d} \left(1 + \log \sigma_{i,j}^2 - \mu_{i,j}^2 - \sigma_{i,j}^2 \right).
\end{align}
This regularization term promotes a smooth, well-structured latent space potentially enhancing the generalization capabilities of model across diverse datasets and task domains.

\subsubsection{Global Attention and Component Weighting}

Following variational encoding, we implement a second-tier adaptive kernel attention mechanism to capture global patterns and interdependencies across the feature landscape as follows:
\begin{align}
\mathbf{G}_i, \mathbf{K}_i^{(glob)}, \mathbf{w}_i^{(glob)} = \text{AdaptiveKernelAttention}(\mathbf{V}_i),
\end{align}
where the adaptive kernel attention mechanism retains the same computational structure as the group-specific attention, but uses a separate parameterization to capture global-scale interactions. This architecture enables the model to capture feature dependencies beyond local group boundaries, forming a richer representation. To capture information at multiple abstraction levels, we extract latent variables via projections:
\begin{align}
\mathbf{T}^{(1)}_i &= \mathbf{G}_i \mathbf{W}^{(1)} \in \mathbb{R}^{c_1}, \\
\mathbf{T}^{(2)}_i &= \mathbf{G}_i \mathbf{W}^{(2)} \in \mathbb{R}^{c_2}, \\
\mathbf{T}^{(3)}_i &= \mathbf{G}_i \mathbf{W}^{(3)} \in \mathbb{R}^{c_3}
\end{align}
where $\mathbf{W}^{(1)}, \mathbf{W}^{(2)}, \mathbf{W}^{(3)} \in \mathbb{R}^{p \times c_k}$ are PLS weight matrices with decreasing dimensions $c_1 > c_2 > c_3$. This hierarchy captures patterns from detailed feature interactions to broader structures.

Within this framework, attention assigns dynamic importance weights to the transformed features, allowing the model to emphasize informative components while reducing noise. This adaptive weighting bridges hierarchical feature extraction and prediction, focusing downstream inference on the most relevant information. Building on the group-specific transformations, the global attention mechanism integrates information across multiple scales, capturing both local and broad feature interactions. This multi-level integration allows the model to effectively represent complex, heterogeneous data while maintaining computational tractability.

\subsubsection{Loss Computation and Feature Importance Score}
Our model is trained with a composite loss function designed to balance accuracy, robustness, and regularization, enabling reliable parameter estimation across varied data conditions. 
\begin{algorithm}[t]
\caption{SCALAR based PLS mechanism}
\label{alg:haka-pls}
\begin{algorithmic}[1]
\State \textbf{Input:} Data matrix $\mathbf{X} \in \mathbb{R}^{n \times p}$, response $\mathbf{y} \in \mathbb{R}^{n \times 1}$, feature groups $\{(s_g, e_g)\}_{g=1}^{G}$
\State \textbf{Output:} Predictions $\hat{\mathbf{y}}$, feature importance scores $\{\hat{I}_j\}_{j=1}^{p}$

\Function{KernelAttention}{$\mathbf{X}_i$}
    \State $\mathbf{K}_i \gets \phi_K(\mathbf{X}_i)$ \Comment{Generate data-dependent kernels}
    \State $\hat{\mathbf{K}}_{i,j} \gets \mathbf{K}_{i,j}/\|\mathbf{K}_{i,j}\|_2$ for $j \in \{1,\ldots,k\}$ 
    \State $\hat{\mathbf{w}}_i \gets \text{softmax}(\phi_W(\mathbf{X}_i))$ \Comment{Compute kernel weights}
    \State $\mathbf{T}_{i,j} \gets \mathbf{X}_i \odot \hat{\mathbf{K}}_{i,j}$ for $j \in \{1,\ldots,k\}$ \Comment{Apply kernels}
    \State $\mathbf{A}_i \gets \sum_{j=1}^{k} \hat{\mathbf{w}}_{i,j} \mathbf{T}_{i,j}$ \Comment{Weighted sum}
    \State $\mathbf{Z}_i \gets \phi_P(\mathbf{A}_i) + \mathbf{X}_i$ \Comment{Apply projection with residual}
    \State \Return $\mathbf{Z}_i, \hat{\mathbf{K}}_i, \hat{\mathbf{w}}_i$
\EndFunction

\State Define $\mathbf{X}^{(g)} = \mathbf{X}_{:, s_g:e_g}$ for each group $g \in \{1,\ldots,G\}$ 

\For{$i = 1$ to $n$}
    \For{$g = 1$ to $G$}
        \State $\mathbf{Z}^{(g)}_i, \mathbf{K}^{(g)}_i, \mathbf{w}^{(g)}_i \gets \Call{KernelAttention}{\mathbf{X}^{(g)}_i}$ 
    \EndFor
    \State $\mathbf{Z}_i \gets [\mathbf{Z}^{(1)}_i, \ldots, \mathbf{Z}^{(G)}_i]$ \Comment{Feature combination}
    
    \State $[\delta_i, \gamma_i] \gets \phi_C(\mathbf{Z}_i)$ where $\delta_i \in [0, 0.4], \gamma_i \in [0.5, 1.0]$ 
    \State $\mathbf{M}_i \sim \text{Bernoulli}(1 - \delta_i)$ \Comment{Random dropout mask}
    \State $\mathbf{S}_i \gets \mathbf{Z}_i + \gamma_i \cdot \phi_T(\mathbf{Z}_i) \odot \mathbf{M}_i / (1 - \delta_i)$ \Statex \Comment{Self-calibration}
    
    \State $[\boldsymbol{\mu}_i, \log \boldsymbol{\sigma}_i] \gets [\phi_{\mu}(\phi_E(\mathbf{S}_i)), \phi_{\sigma}(\phi_E(\mathbf{S}_i))]$ \Statex\Comment{ Variational parameters}
    \State $\mathbf{V}_i \gets \mathbf{S}_i + \phi_D(\boldsymbol{\mu}_i + \boldsymbol{\epsilon}_i \odot \exp(\log \boldsymbol{\sigma}_i / 2))$ 
    \Statex \hspace{0.6cm}where $\boldsymbol{\epsilon}_i \sim \mathcal{N}(0, \mathbf{I})$ \Comment{VAE encoding}
    
    \State $\mathbf{Z}_i^{(glob)}, \mathbf{K}_i^{(glob)}, \mathbf{w}_i^{(glob)} \gets \Call{KernelAttention}{\mathbf{V}_i}$ \Statex \Comment{Global attention}
    
    \State $\mathbf{T}^{(k)}_i \gets \mathbf{Z}_i^{(glob)} \mathbf{W}^{(k)}, \forall k =1,2,3$ \Comment{PLS components}
    \State $\hat{y}_i \gets \phi_y\left([\mathbf{T}^{(1)}_i, \mathbf{T}^{(2)}_i, \mathbf{T}^{(3)}_i] \odot \text{softmax}(\phi_{\alpha}(\mathbf{Z}_i^{(glob)}))\right)$ 
\EndFor

\State $
\mathcal{L} \gets \omega_{\text{MSE}} \cdot \mathcal{L}_{\text{MSE}} 
+ (1 - \omega_{\text{MSE}}) \cdot \mathcal{L}_{\text{Huber}}  $

\Statex $\quad\quad+ \min(1.0, \frac{\text{epoch}}{\text{total\_epochs} \cdot 0.1}) \cdot \beta_0 \cdot \mathcal{L}_{\text{KL}} 
$
\State $\hat{I}_j \gets \frac{\frac{1}{n} \sum_{i=1}^{n} |\sum_{l=1}^{k} w_{i,l}^{(glob)} \cdot K_{i,l,j}^{(glob)}| - \min_j I_j}{\max_j I_j - \min_j I_j}$ for $j \in \{1,\ldots,p\}$ \Comment{Feature importance}

\State \Return $\{\hat{y}_i\}_{i=1}^{n}, \{\hat{I}_j\}_{j=1}^{p}$
\end{algorithmic}
\label{algo1}
\end{algorithm}

The core loss is the mean squared error (MSE), which drives precise prediction of the target variable in a supervised regression setting. To improve robustness against outliers and noise, we integrate a dynamically weighted Huber loss that smoothly transitions between squared and absolute error based on an adaptive threshold. To ensure stable and informative latent representations, we include Kullback-Leibler (KL) divergence regularization. This term prevents collapse of the learned feature distributions and its influence is gradually adjusted during training, facilitating a smooth balance between representation learning and predictive performance. 
The overall training objective can be written as a weighted sum:
\begin{align*}
\mathcal{L} &= \omega_{\text{MSE}} \mathcal{L}_{\text{MSE}} + (1 - \omega_{\text{MSE}}) \mathcal{L}_{\text{Huber}} + \min\left(1, \dfrac{t}{0.1}\right) \beta_0 \mathcal{L}_{\text{KL}},
\end{align*}
where $t = ({\text{epoch}}/{\text{total\_epochs}})$ and the loss components are defined as:
\begin{align}
\mathcal{L}_{\text{MSE}} &= \frac{1}{n} \sum_{i=1}^{n} (y_i - \hat{y}_i)^2, \\\mathcal{L}_{\text{Huber}} &= \frac{1}{n} \sum_{i=1}^{n} 
\begin{cases} 
\frac{1}{2}(y_i - \hat{y}_i)^2 & \text{for } |y_i - \hat{y}_i| \leq \delta \\
\delta(|y_i - \hat{y}_i| - \frac{1}{2}\delta) & \text{otherwise}
\end{cases},
\end{align}
and $\mathcal{L}_{\text{KL}}$ is defined is \ref{eq10}. The complete algorithm is detailed in Algorithm \ref{algo1}.
To offer insights into model decisions, we perform post hoc analysis to estimate feature importance derived from the PLS components. These importance scores reflect the contributions of individual features within the adaptive kernel attention framework. The scores are subsequently normalized to place them on a consistent scale, enabling meaningful comparison across features:
\begin{align}
I_j &= \frac{1}{n} \sum_{i=1}^{n} \left| \sum_{l=1}^{k} w_{i,l}^{(glob)} \cdot K_{i,l,j}^{(glob)} \right|, \\\hat{I}_j &= \frac{I_j - \min_j I_j}{\max_j I_j - \min_j I_j}
\end{align}
where \(I_j\) represents the importance score for feature \(j\).
Our framework combines attention weights and kernel activations to assign quantitative relevance scores to individual features. While these scores do not fully explain model decisions, they provide a meaningful indication of feature influence within the predictive process. 

\section{Experimental Results}
In this section, we evaluate the performance of our proposed SCALAR architecture against state-of-the-art methods on two different datasets. All experiments use PyTorch 2.1.2 with CUDA 11.2, GPU Tesla T4, 15GB RAM, and Mac OS. To conserve space and maintain readability, additional experimental results are included in the Appendix (supplementary material). The implementation code to reproduce the experimental results can be found at: https://anonymous.4open.science/r/SCALAR-CF01.

\subsection{Results on Davis Dataset}
\begin{figure}[t]
    \centering
    \includegraphics[width=0.9\linewidth]{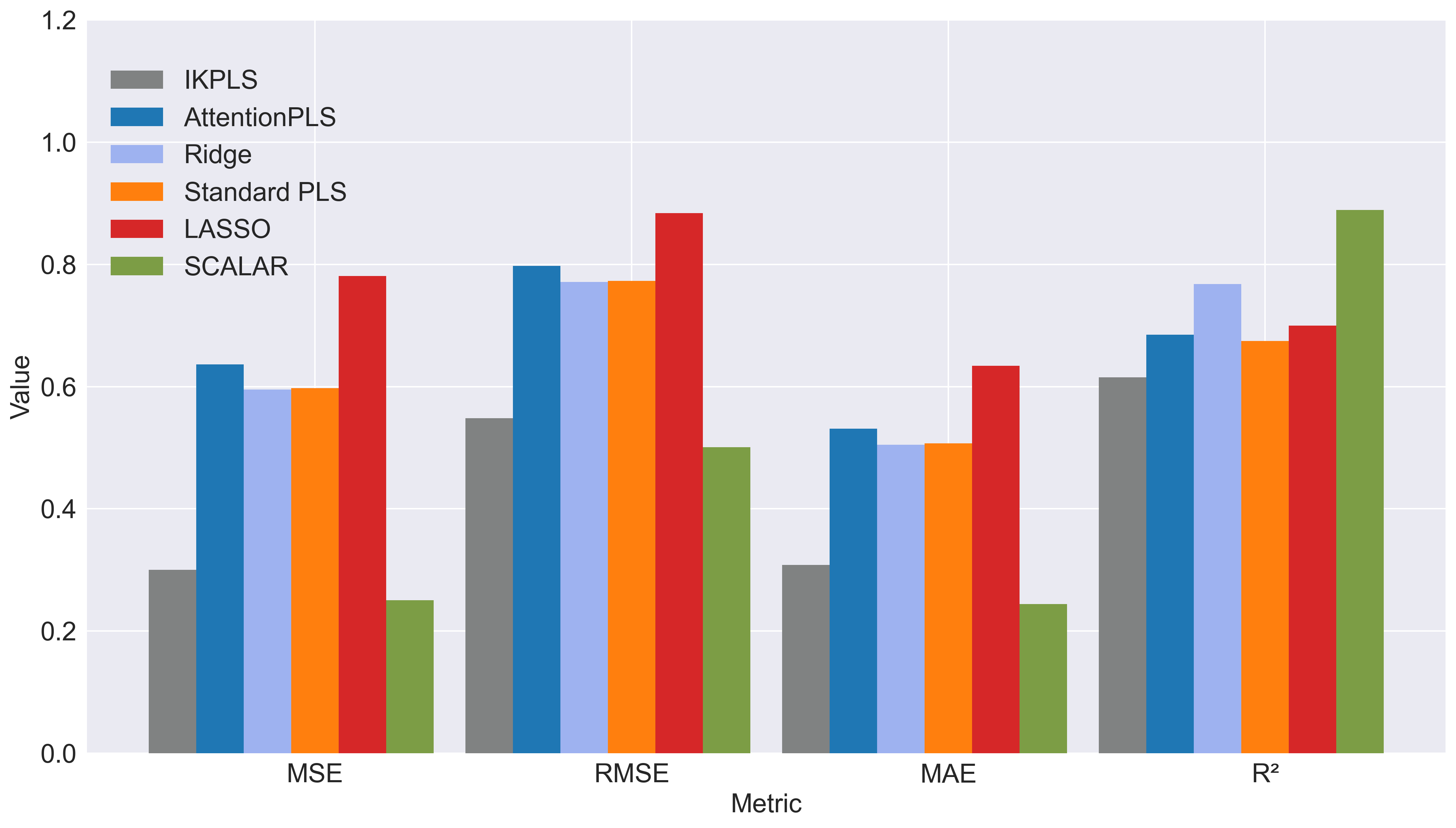}
    \caption{Comparative Analysis for different metric values.}
    \label{fig:compare_scalar}
\end{figure}

\begin{figure}[t]
\centering
\includegraphics[width=0.9\linewidth]{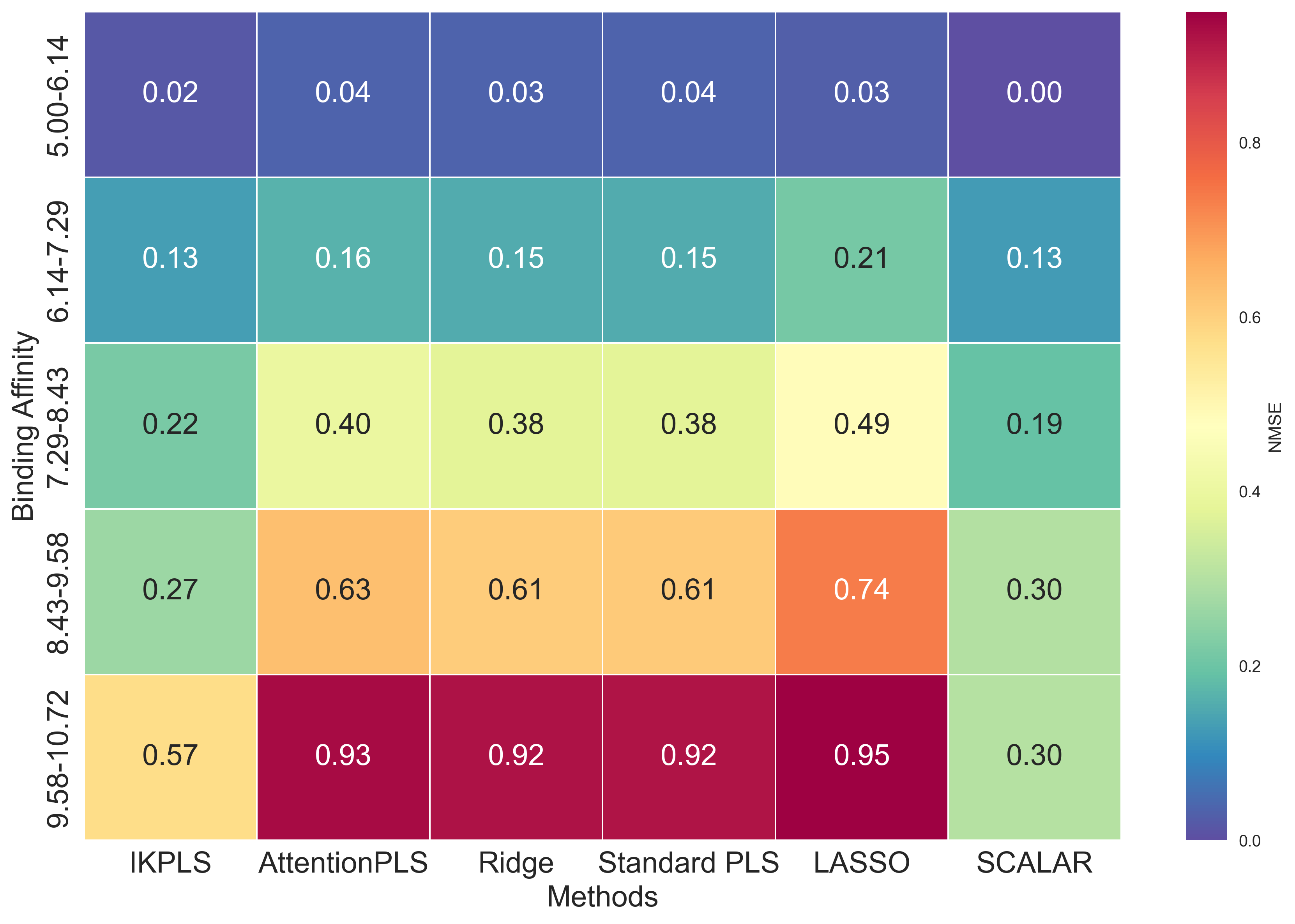}
\caption{Bin-wise RMSE for different methods.}
\label{fig:bin}
\vspace{-1\baselineskip}
\end{figure}

\begin{figure}[t]
    \centering
    \includegraphics[width=0.9\linewidth]{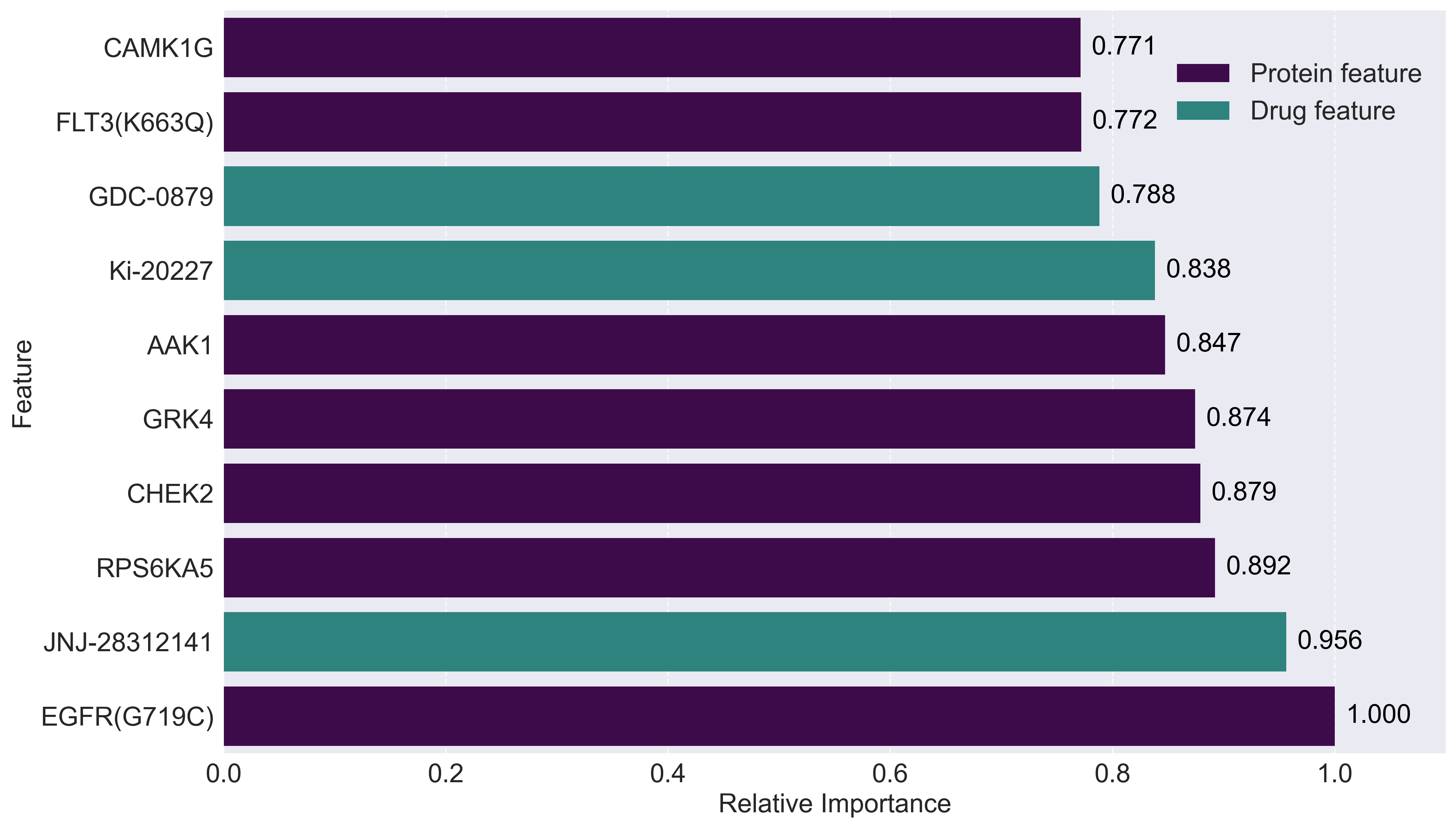} 
    \caption{Feature importance estimated by kernel based attention}
    \label{fig:imp}
\end{figure}

\begin{figure}[t]
    \centering
    \includegraphics[width=0.9\linewidth]{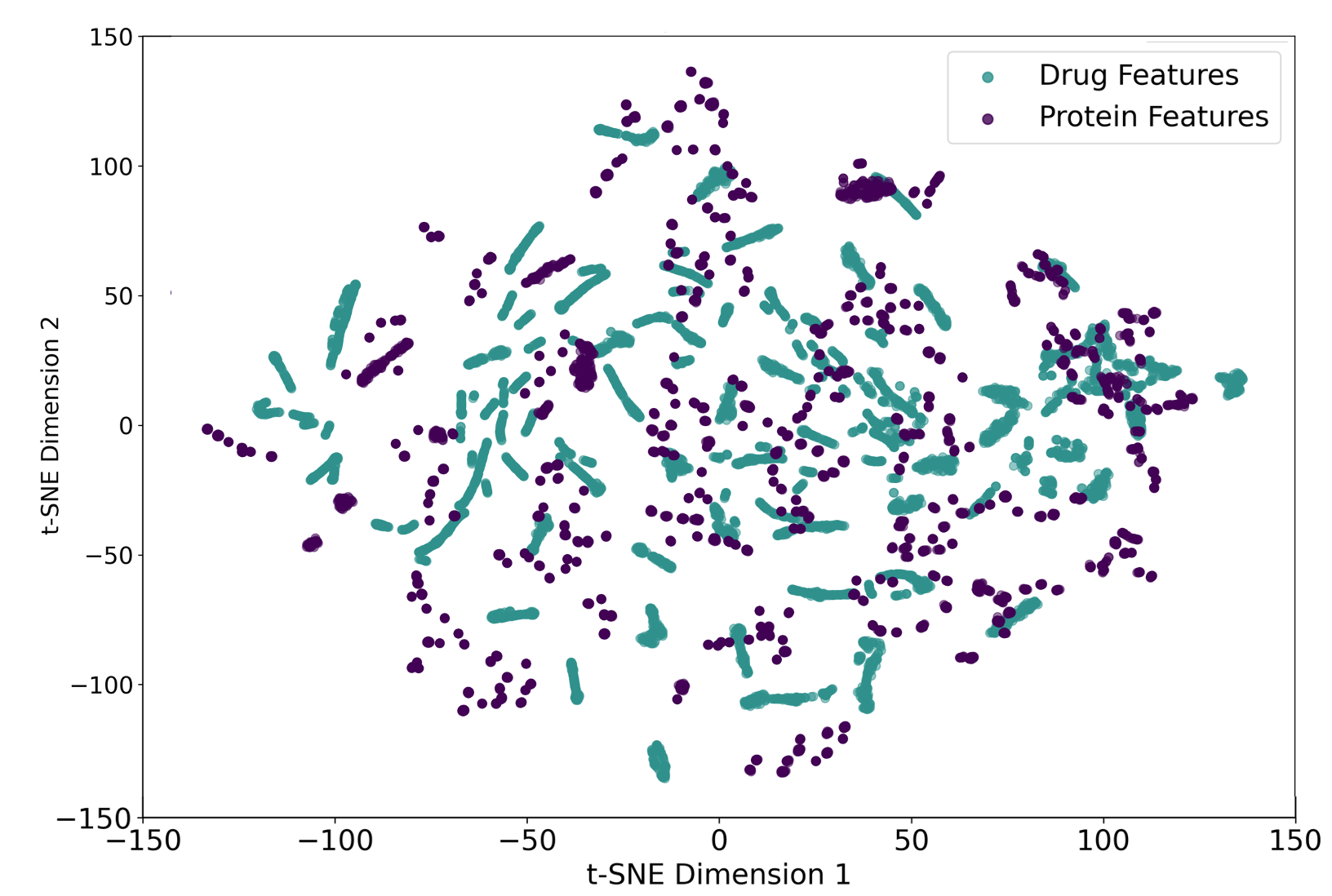}
    \caption{Distribution of feature groups in latent space.}
    \label{fig:tsne}
    \vspace{-1\baselineskip}
\end{figure}

We assessed the regression performance on the relatively complex Davis binding affinity dataset \cite{davis2011comprehensive} in which the prediction task is to determine continuous values that indicate the strength of the binding between the drug and the target. First we preprocessed the data by converting all features to numeric representations. For molecular structures, we utilized RDKit \cite{landrum2013rdkit} to generate Morgan fingerprints \cite{martin2021state} and key descriptors, applying PCA to reduce dimensionality while preserving variance. For protein sequences, we computed normalized amino acid composition frequencies and dipeptide patterns. The comparative performance on regression is demonstrated in Figure~\ref{fig:compare_scalar}. In the top figure, our proposed method SCALAR shows superior performance across all key metrics. In Figure~\ref{fig:bin}, the root mean squared error values for different bins of binding affinity are compared to demonstrate generalizability, particularly in cases with fewer samples. For example, in the 9.58-10.72 affinity bin, there are only 14 samples. Despite this, SCALAR is able to learn patterns from these limited samples, whereas other methods fail to do so. This addresses the data imbalance challenge in kinase binding datasets, demonstrating robust generalizability without overfitting to moderate-affinity bins.
Figure~\ref{fig:imp} provides insights into feature importance values computed in Algorithm \ref{algo1}. EGFR(D719C) has been identified as the most influential feature (1.0), followed by JNJ-28312141 (0.956) and RPS6KA3 (0.892). Our top features include a mix of kinases (like EGFR, RPS6KA5, CHEK2) and drugs (JNJ-28312141, Ki-20227, GDC-0879), which is reasonable for a binding affinity prediction task. The presence of clinically relevant kinases (like EGFR and FLT3) in top features is consistent with established biological knowledge, supporting the biological relevance of the feature selection of model. The t-SNE visualization in Figure \ref{fig:tsne} reveals distinct organizational patterns between drug and protein features in the latent space.

Protein features form scattered, point-like clusters distributed throughout the space, while drug features create elongated, linear arrangements that form path-like structures. 
This organization indicates fundamental differences in information encoding, i.e. protein features likely reflect diverse kinase structural properties with distinct functional groups, while drug features suggest shared pharmacophore patterns that follow more continuous variations. The distinct yet complementary distribution pattern demonstrates the ability of model to capture both individual feature types and their inherent structural relationships, while also preserving the global interactions between features.

\begin{figure}[t]
    \centering
    \includegraphics[width=0.9\linewidth]{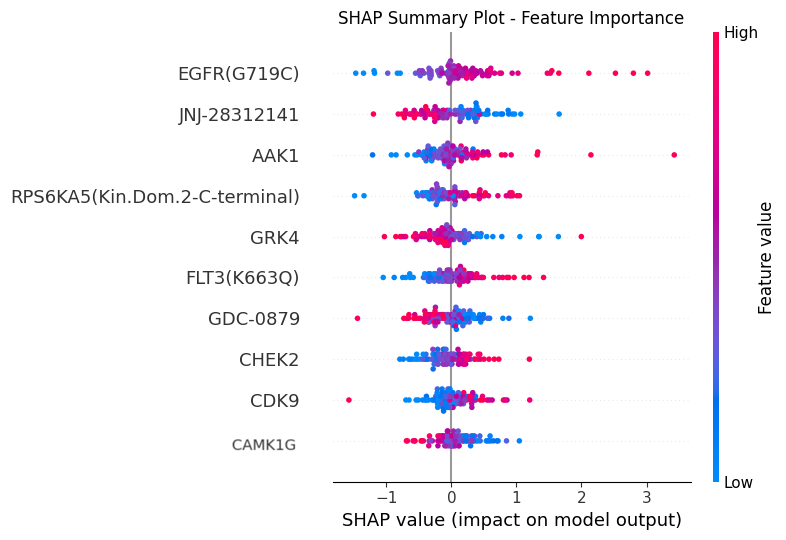} 
    \caption{SHAP summary plot showing most influetial features in binding affinity prediction task.}
    \label{fig:shap1}
\end{figure}

\begin{figure}[t]
    \centering
    \includegraphics[width=0.9\linewidth]{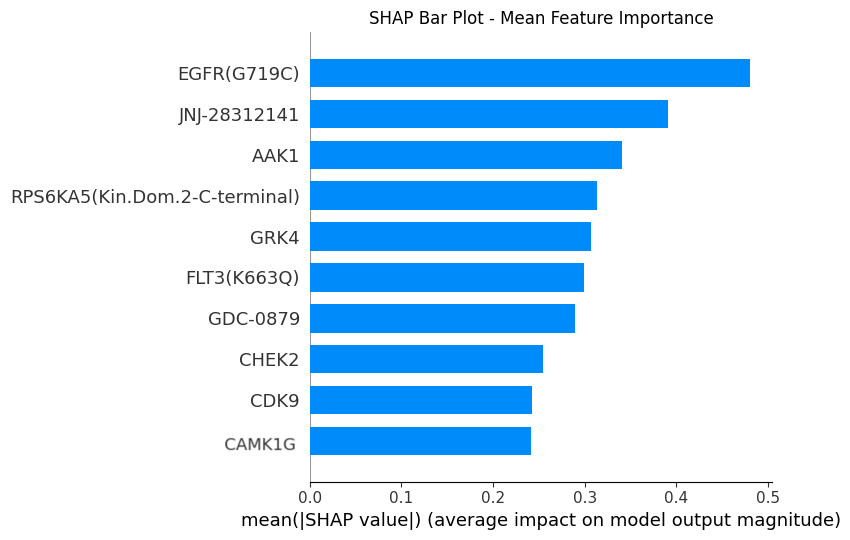} 
    \caption{Average SHAP values for different drug and kinase features}
    \label{fig:shap2}
    \vspace{-1\baselineskip}
\end{figure}

The SHAP analysis in Figure \ref{fig:shap1} and Figure \ref{fig:shap2} further supported the feature importance rankings identified by the SCALAR model, highlighting key contributors to the binding affinity predictions. Notably, EGFR(D719C) emerged as the most influential feature, exhibiting the highest average SHAP value, thereby underscoring its strong predictive relevance. This was closely followed by the small-molecule inhibitor JNJ-28312141 (0.956) and the kinase RPS6KA3 (0.892), both of which demonstrated substantial contributions across the SHAP summary and average SHAP value plots. Importantly, several top-ranking features comprised a biologically coherent mix of kinases including AAK1, RPS6KA5, and GRK4. The pharmacological compounds such GDC-0879 has been identified as an influential feature by both methods in Figure \ref{fig:imp} and Figure \ref{fig:shap1}. The presence of clinically validated kinases like EGFR and FLT3 further supports the biological relevance and interpretability of the model. The SHAP summary plot revealed consistent patterns of feature influence, with the magnitude and direction of individual SHAP values aligning with the SCALAR-derived feature importance scores. These converging lines of evidence confirm that the selected features are consistently important and biologically meaningful.

The insights provided by the SHAP analysis not only validate the model’s decision-making process but also offer valuable interpretability that can inform drug design optimization. By quantitatively attributing feature contributions to binding affinity predictions, SHAP helps identify key kinase mutations and drug-target interactions driving model outputs. For instance, the high SHAP values associated with clinically relevant kinases such as EGFR(D719C) and FLT3, as well as potent inhibitors like JNJ-28312141, underscore their critical roles in modulating binding affinity. Such findings can inform the prioritization of molecular targets and compound modifications in the early stages of drug development, suggesting that integrating explainable machine learning methods like SHAP into predictive modeling pipelines enhances both the accuracy and actionable insight of computational drug discovery workflows.

Table~\ref{tab:sota_comparison} presents a comparative evaluation of our proposed model, SCALAR, against several established state-of-the-art deep learning models for drug–target interaction (DTI) prediction on the Davis dataset. Notably, SCALAR achieves a concordance index (CI) of 0.901, outperforming models such as DeepDTA (0.878), WideDTA (0.886), and GraphDTA (0.881). In terms of mean squared error (MSE), SCALAR matches the lowest reported value (0.245), indicating strong predictive accuracy and robustness. These results demonstrate that SCALAR not only improves predictive performance but also competes favorably with leading models, confirming its effectiveness for binding affinity prediction tasks.

\begin{table}[t]
\centering
\caption{Performance comparison of SCALAR with state-of-the-art DTI prediction models on the Davis dataset.}
\begin{tabular*}{\columnwidth}{@{\extracolsep{\fill}}lcc}
\toprule
\textbf{Model} & \textbf{Concordance Index} & \textbf{MSE} \\
\midrule
DeepDTA \cite{ozturk2018deepdta}      & 0.878 & 0.261 \\
WideDTA \cite{ozturk2019widedta}      & 0.886 & 0.262 \\
GraphDTA \cite{nguyen2021graphdta}    & 0.881 & \textbf{0.245} \\
SCALAR                & \textbf{0.901} & 0.256 \\
\bottomrule
\end{tabular*}
\label{tab:sota_comparison}
\vspace{-1\baselineskip}
\end{table}

\subsection{Results on NIR Spectroscopy for Wheat Dataset}
We also evaluated the regression performance on the wheat kernels dataset \cite{pedersen2002near}, comprising 415 calibration and 108 validation samples across 100 wavelengths (850-1050 nm) to show comparison with Attention-PLS \cite{xiong2022soft}. The target variable was protein content. Using 80/20 train/test split with 10-fold cross-validation, we selected optimal latent components via explained variance criterion. The predictive performance of several methods has been compared in Table \ref{tab:performance_comparison}.

\begin{table}[htbp]
\centering
\setlength{\tabcolsep}{12pt}
\renewcommand{\arraystretch}{1.2}
\caption{Performance Comparison of Different Methods on Wheat Data}
\begin{tabular*}{\columnwidth}{@{\extracolsep{\fill}}cccc@{}}
\hline
\textbf{Method} & \textbf{RMSE} & \textbf{MAE} & $\mathbf{R^2}$ \\
\hline
Ridge & 0.7125 & 0.8125 & 0.7510 \\
LASSO & 0.6500 & 0.6800 & 0.6600 \\
PLS & 0.6106 & 0.4618 & 0.7930 \\
IKPLS \cite{briscik2024supervised} & 0.1060 & 0.7618 & 0.8130 \\
AttentionPLS \cite{xiong2022soft} & 0.0240 & 0.0118 & 0.8280 \\
SCALAR & \textbf{0.0141} & \textbf{0.0016} & \textbf{0.8930} \\
\hline
\end{tabular*}
\label{tab:performance_comparison}
\end{table}

Table~\ref{tab:performance_comparison} shows SCALAR outperforming all other approaches with the lowest RMSE (0.0141) and highest $R^2$ (0.8930). Attention-PLS \cite{xiong2022soft} also performs well ($R^2$=0.8280), while conventional methods show notably higher error rates. These results demonstrate advanced methods like SCALAR enhance predictive accuracy over traditional techniques.

\section{Ablation Study}
We examined the role of various components in SCALAR from three different perspectives: robustness to domain shifts, generalization with limited data, and computational efficiency.

\subsection{Self-calibration under domain shifts:} Removing self-calibration (SCALAR w/o SC) caused clear degradation. Within-domain RMSE rose moderately (0.0141→0.0168, 19.1\%), but under cross-domain shifts error nearly doubled (0.0392→0.0751, 91.6\%). Adversarial conditions showed the strongest effect (0.0587→0.1284, 118.7\%). Thus, self-calibration is especially critical for stability under distributional changes.

\subsection{Variational encoding for generalization:} Comparing SCALAR with and without variational encoding revealed increasing benefits as data size decreased. With full training data, the gain in $R^2$ was modest (0.893→0.871, 2.5\%), but at 10\% data it became substantial (0.732→0.521, 28.8\%). Hence, it indicates that variational encoding mitigates overfitting in scarce-data regimes.

\subsection{Computational efficiency:} SCALAR requires more resources than simpler baselines: training takes 318s versus 167s for AttentionPLS, and inference 4.1s versus 3.4s. Nevertheless, the overhead is modest relative to the accuracy gains, and inference remains efficient for real-time use. The results affirm that each component contributes critically, and together they make SCALAR a robust modeling framework.

\section{Discussion}
The SCALAR model presents a significant advancement over conventional dimensionality reduction and regression techniques, particularly for high-dimensional, heterogeneous data with complex feature interactions. By integrating adaptive kernel attention and group-specific processing, SCALAR effectively captures local and global dependencies. Experimental results on the Davis dataset and Wheat NIR spectroscopy dataset demonstrate superior predictive accuracy ($R^2$) and reduced error metrics (e.g., RMSE, MAE) compared to established methods such as PLS and AttentionPLS. Adaptive mechanisms, particularly self-calibration and dynamic loss weighting, enhance robustness in scenarios with limited samples or high noise levels.

While effective, the architectural complexity of SCALAR increases computational demands and necessitates careful hyperparameter tuning, particularly for adaptive and variational components.  Additionally, while the model has shown promising results, further validation across a wider range of tasks is necessary to assess its generalizability. Future work will focus on algorithmic efficiency improvements and validation across additional domains to establish broader applicability while maintaining performance gain.

\section{Conclusion}
This paper introduces SCALAR, a novel architecture that bridges the gap between predictive performance and interpretability in high-dimensional, heterogeneous data environments. By integrating adaptive kernel-based attention with a self-calibration layer, the model dynamically adjusts to contextual variations, capturing both local feature interactions and global dependencies with greater precision than traditional PLS methods. Unlike conventional approaches, this framework assigns variable significance to distinct feature groups through adaptive kernel functions.
The incorporation of a variational latent space with dual Kullback-Leibler divergence regularization enhances generalization while controlling representational complexity. SCALAR demonstrates unique capabilities in disentangling scale-specific interactions, revealing context-dependent patterns typically masked in traditional modeling. The multi-level attention mechanism improves interpretability by facilitating effective feature selection without introducing unnecessary complexity.
Empirical evaluations confirm SCALAR significantly outperforms conventional methods in predictive accuracy and error reduction. The self-calibration mechanism strengthens robustness by mitigating overfitting and promoting cross-domain generalizability. These findings underscore the broad applicability of our model across fields requiring both interpretability and predictive power.

\bibliographystyle{unsrt}
\bibliography{references}

\begin{thebibliography}{10}

\bibitem{jia2022feature}
Weikuan Jia, Meili Sun, Jian Lian, and Sujuan Hou.
\newblock Feature dimensionality reduction: a review.
\newblock {\em Complex \& Intelligent Systems}, 8(3):2663--2693, 2022.

\bibitem{montesinos2022overfitting}
Osval~Antonio Montesinos~L{\'o}pez, Abelardo Montesinos~L{\'o}pez, and Jose Crossa.
\newblock Overfitting, model tuning, and evaluation of prediction performance.
\newblock In {\em Multivariate statistical machine learning methods for genomic prediction}, pages 109--139. Springer, 2022.

\bibitem{vogelstein2021supervised}
Joshua~T Vogelstein, Eric~W Bridgeford, Minh Tang, Da~Zheng, Christopher Douville, Randal Burns, and Mauro Maggioni.
\newblock Supervised dimensionality reduction for big data.
\newblock {\em Nature communications}, 12(1):2872, 2021.

\bibitem{goel2024machine}
Diksha Goel, Hussain Ahmad, Ankit~Kumar Jain, and Nikhil~Kumar Goel.
\newblock Machine learning driven smishing detection framework for mobile security.
\newblock {\em arXiv preprint arXiv:2412.09641}, 2024.

\bibitem{chopra2024chatnvd}
Shivansh Chopra, Hussain Ahmad, Diksha Goel, and Claudia Szabo.
\newblock Chatnvd: Advancing cybersecurity vulnerability assessment with large language models.
\newblock {\em arXiv preprint arXiv:2412.04756}, 2024.

\bibitem{ahmad2025survey}
Hussain Ahmad, Faheem Ullah, and Rehan Jafri.
\newblock A survey on immersive cyber situational awareness systems.
\newblock {\em Journal of Cybersecurity and Privacy}, 5(2):33, 2025.

\bibitem{ullah2025skills}
Faheem Ullah, Xiaohan Ye, Uswa Fatima, Zahid Akhtar, Yuxi Wu, and Hussain Ahmad.
\newblock What skills do cyber security professionals need?
\newblock {\em arXiv preprint arXiv:2502.13658}, 2025.

\bibitem{ahmad2025future}
Hussain Ahmad and Diksha Goel.
\newblock The future of ai: Exploring the potential of large concept models.
\newblock {\em arXiv:2501.05487}, 2025.

\bibitem{rosipal2001kernel}
Roman Rosipal and Leonard~J Trejo.
\newblock Kernel partial least squares regression in reproducing kernel hilbert space.
\newblock {\em Journal of machine learning research}, 2(Dec):97--123, 2001.

\bibitem{mcinnes2020umapuniformmanifoldapproximation}
Leland McInnes, John Healy, and James Melville.
\newblock Umap: Uniform manifold approximation and projection for dimension reduction, 2020.

\bibitem{kobak2021initialization}
Dmitry Kobak and George~C Linderman.
\newblock Initialization is critical for preserving global data structure in both t-sne and umap.
\newblock {\em Nature biotechnology}, 39(2):156--157, 2021.

\bibitem{bengio2013representation}
Yoshua Bengio, Aaron Courville, and Pascal Vincent.
\newblock Representation learning: A review and new perspectives.
\newblock {\em IEEE transactions on pattern analysis and machine intelligence}, 35(8):1798--1828, 2013.

\bibitem{xu2019understanding}
Jingjing Xu, Xu~Sun, Zhiyuan Zhang, Guangxiang Zhao, and Junyang Lin.
\newblock Understanding and improving layer normalization.
\newblock {\em Advances in neural information processing systems}, 32, 2019.

\bibitem{abbas2024robust}
Farwa Abbas and Hussain Ahmad.
\newblock Robust partial least squares using low rank and sparse decomposition.
\newblock {\em arXiv:2407.06936}, 2024.

\bibitem{arik2021tabnet}
Sercan~{\"O} Arik and Tomas Pfister.
\newblock Tabnet: Attentive interpretable tabular learning.
\newblock In {\em Proceedings of the AAAI conference on artificial intelligence}, volume~35, pages 6679--6687, 2021.

\bibitem{jain2019attention}
Sarthak Jain and Byron~C Wallace.
\newblock Attention is not explanation.
\newblock {\em arXiv preprint arXiv:1902.10186}, 2019.

\bibitem{kumar2021partial}
Keshav Kumar.
\newblock Partial least square (pls) analysis: Most favorite tool in chemometrics to build a calibration model.
\newblock {\em Resonance}, 26:429--442, 2021.

\bibitem{anwardeen2023statistical}
Najeha~R Anwardeen, Ilhame Diboun, Younes Mokrab, Asma~A Althani, and Mohamed~A Elrayess.
\newblock Statistical methods and resources for biomarker discovery using metabolomics.
\newblock {\em BMC bioinformatics}, 24(1):250, 2023.

\bibitem{mihalik2022canonical}
Agoston Mihalik, James Chapman, Rick~A Adams, Nils~R Winter, Fabio~S Ferreira, John Shawe-Taylor, Janaina Mour{\~a}o-Miranda, Alzheimer’s Disease~Neuroimaging Initiative, et~al.
\newblock Canonical correlation analysis and partial least squares for identifying brain--behavior associations: A tutorial and a comparative study.
\newblock {\em Biological Psychiatry: Cognitive Neuroscience and Neuroimaging}, 7(11):1055--1067, 2022.

\bibitem{engstrom2024ikpls}
Ole-Christian~Galbo Engstr{\o}m, Erik~Schou Dreier, Birthe~M{\o}ller Jespersen, and Kim~Steenstrup Pedersen.
\newblock Ikpls: Improved kernel partial least squares and fast cross-validation algorithms for python with cpu and gpu implementations using numpy and jax.
\newblock {\em Journal of Open Source Software}, 9(99):6533, 2024.

\bibitem{briscik2024supervised}
Mitja Briscik, Gabriele Tazza, L{\'a}szl{\'o} Vid{\'a}cs, Marie-Agn{\`e}s Dillies, and S{\'e}bastien D{\'e}jean.
\newblock Supervised multiple kernel learning approaches for multi-omics data integration.
\newblock {\em BioData Mining}, 17(1):53, 2024.

\bibitem{kong2022deep}
Xiangyin Kong and Zhiqiang Ge.
\newblock Deep pls: A lightweight deep learning model for interpretable and efficient data analytics.
\newblock {\em IEEE transactions on neural networks and learning systems}, 34(11):8923--8937, 2022.

\bibitem{ozturk2018deepdta}
Hakime {\"O}zt{\"u}rk, Arzucan {\"O}zg{\"u}r, and Elif Ozkirimli.
\newblock Deepdta: deep drug–target binding affinity prediction.
\newblock {\em Bioinformatics}, 34(17):i821--i829, 2018.

\bibitem{lee2019deepconv}
Ingoo Lee, Jongsoo Keum, and Hojung Nam.
\newblock Deepconv-dti: Prediction of drug-target interactions via deep learning with convolution on protein sequences.
\newblock {\em PLoS computational biology}, 15(6):e1007129, 2019.

\bibitem{lee2024dlm}
Jonghyun Lee, Dae~Won Jun, Ildae Song, and Yun Kim.
\newblock Dlm-dti: a dual language model for the prediction of drug-target interaction with hint-based learning.
\newblock {\em Journal of Cheminformatics}, 16(1):14, 2024.

\bibitem{proteinLigandBinding2024}
Huiwen Wang.
\newblock Prediction of protein-ligand binding affinity via deep learning models.
\newblock {\em Briefings in bioinformatics}, 25,2, 2024.

\bibitem{10849366}
Lingtao Chen, Kazi~Fahim Ahmad~Nasif, Bobin Deng, Shuteng Niu, and Chloe~Yixin Xie.
\newblock Predicting protein-protein binding affinity with deep learning: A comparative analysis of cnn and transformer models.
\newblock In {\em 2024 IEEE 36th International Conference on Tools with Artificial Intelligence (ICTAI)}, pages 548--555, 2024.

\bibitem{xiong2022soft}
Yinran Xiong, Wuye Yang, Huiyun Liao, Zhenlin Gong, Zhenzhen Xu, Yiping Du, and Wei Li.
\newblock Soft variable selection combining partial least squares and attention mechanism for multivariable calibration.
\newblock {\em Chemometrics and Intelligent Laboratory Systems}, 223:104532, 2022.

\bibitem{hu2018squeeze}
Jie Hu, Li~Shen, and Gang Sun.
\newblock Squeeze-and-excitation networks.
\newblock In {\em Proceedings of the IEEE conference on computer vision and pattern recognition}, pages 7132--7141, 2018.

\bibitem{chen2020dynamic}
Yinpeng Chen, Xiyang Dai, Mengchen Liu, Dongdong Chen, Lu~Yuan, and Zicheng Liu.
\newblock Dynamic convolution: Attention over convolution kernels.
\newblock In {\em Proceedings of the IEEE/CVF conference on computer vision and pattern recognition}, pages 11030--11039, 2020.

\bibitem{kadra2021well}
Arlind Kadra, Marius Lindauer, Frank Hutter, and Josif Grabocka.
\newblock Well-tuned simple nets excel on tabular datasets.
\newblock {\em Advances in neural information processing systems}, 34:23928--23941, 2021.

\bibitem{song2019autoint}
Weiping Song, Chence Shi, Zhiping Xiao, Zhijian Duan, Yewen Xu, Ming Zhang, and Jian Tang.
\newblock Autoint: Automatic feature interaction learning via self-attentive neural networks.
\newblock In {\em Proceedings of the 28th ACM international conference on information and knowledge management}, pages 1161--1170, 2019.

\bibitem{sonderby2016ladder}
Casper~Kaae S{\o}nderby, Tapani Raiko, Lars Maal{\o}e, S{\o}ren~Kaae S{\o}nderby, and Ole Winther.
\newblock Ladder variational autoencoders.
\newblock {\em Advances in neural information processing systems}, 29, 2016.

\bibitem{vidaurre2013bayesian}
Diego Vidaurre, Marcel~AJ Van~Gerven, Concha Bielza, Pedro Larranaga, and Tom Heskes.
\newblock Bayesian sparse partial least squares.
\newblock {\em Neural computation}, 25(12):3318--3339, 2013.

\bibitem{davis2011comprehensive}
Mindy~I Davis, Jeremy~P Hunt, Sanna Herrgard, Pietro Ciceri, Lisa~M Wodicka, Gabriel Pallares, Michael Hocker, Daniel~K Treiber, and Patrick~P Zarrinkar.
\newblock Comprehensive analysis of kinase inhibitor selectivity.
\newblock {\em Nature biotechnology}, 29(11):1046--1051, 2011.

\bibitem{landrum2013rdkit}
Greg Landrum.
\newblock Rdkit documentation.
\newblock {\em Release}, 1(1-79):4, 2013.

\bibitem{martin2021state}
Lewis Martin.
\newblock State of the art iterative docking with logistic regression and morgan fingerprints.
\newblock 2021.

\bibitem{ozturk2019widedta}
Hakime {\"O}zt{\"u}rk, Arzucan {\"O}zg{\"u}r, and Elif Ozkirimli.
\newblock Widedta: prediction of drug–target binding affinity.
\newblock {\em arXiv preprint arXiv:1902.04166}, 2019.

\bibitem{nguyen2021graphdta}
Tuan Nguyen, Thang Le, Terrence Quinn, Thin Nguyen, and Svetha Venkatesh.
\newblock Graphdta: Predicting drug-target binding affinity with graph neural networks.
\newblock In {\em Bioinformatics}, volume~37, pages i114--i121. Oxford University Press, 2021.

\bibitem{pedersen2002near}
Dorthe~Kj{\ae}r Pedersen, Harald Martens, Jesper~Pram Nielsen, and S{\o}ren~Balling Engelsen.
\newblock Near-infrared absorption and scattering separated by extended inverted signal correction (eisc): Analysis of near-infrared transmittance spectra of single wheat seeds.
\newblock {\em Applied spectroscopy}, 56(9):1206--1214, 2002.

\end{thebibliography}
\end{document}